\documentclass[journal]{IEEEtran}
\usepackage[dvipsnames]{xcolor}
\usepackage{fancyvrb}
\usepackage{caption}
\usepackage{subcaption}
\usepackage{enumitem}
\usepackage{amsmath}
\definecolor{mypink1}{rgb}{0.858, 0.188, 0.478}
\definecolor{mypink2}{RGB}{219, 48, 122}

\ifCLASSINFOpdf
\else
\fi

\usepackage[dvipsnames]{xcolor}
\usepackage{fancyvrb}
\usepackage{caption}
\usepackage{subcaption}
\usepackage{enumitem}
\usepackage{amsmath}

\usepackage{tikz}
\usepackage{algorithmic}
\usepackage{graphicx}
\usepackage{caption}
\usepackage{algorithm2e}
\usepackage{amssymb}
\usepackage{amsmath}
\usepackage{xcolor}
\def\BibTeX{{\rm B\kern-.05em{\sc i\kern-.025em b}\kern-.08em
    T\kern-.1667em\lower.7ex\hbox{E}\kern-.125emX}}

\begin{document}
%
\title{{Hybrid Dynamic Pruning: A Pathway to Efficient Transformer Inference}}
%
%
%

\author{Ghadeer~A.~Jaradat,~\IEEEmembership{}
        Mohammed~F.~Tolba,
        Ghada~Alsahli, Hani~Saleh,~\IEEEmembership{Member, IEEE},
    Mahmoud~Al-Qutayri,~\IEEEmembership{Member, IEEE},
     Thanos~Stouraitis,~\IEEEmembership{ Life Fellow, IEEE},
    Baker~Mohammad,~\IEEEmembership{Member, IEEE},
}

\maketitle 

\begin{abstract}
In the world of deep learning, Transformer models have become very significant, leading to improvements in many areas from understanding language to recognizing images, covering a wide range of applications. Despite their success, the deployment of these models in real-time applications, particularly on edge devices, poses significant challenges due to their quadratic computational intensity and memory demands. To overcome these challenges we introduce a novel Hybrid Dynamic Pruning (HDP), an efficient algorithm-architecture co-design approach that accelerates transformers using head sparsity, block sparsity and approximation opportunities to reduce computations in attention and reduce memory access. With the observation of the huge redundancy in attention scores and attention heads, we propose a novel integer-based row-balanced block pruning to prune unimportant blocks in the attention matrix at run time, also propose integer-based head pruning to detect and prune unimportant heads at an early stage at run time. Also we propose an approximation method that reduces attention computations. To efficiently support these methods with lower latency and power efficiency, we  propose a HDP co-processor architecture.

\end{abstract}

\begin{IEEEkeywords}
Hardware acceleration, dynamic pruning, approximation, self attention, acceleration.
\end{IEEEkeywords}

%
\IEEEpeerreviewmaketitle

\section{Introduction}
Transformer models, including BERT\cite{devlin2018bert}, GPT \cite{radford2018improving}, T5 \cite{raffel2020exploring} and others \cite{liu2019roberta} \cite{lewis2019bart} , have transformed Natural Language Processing (NLP) with their attention mechanism, achieving top performance in tasks such as question-answering \cite{radford2019language}, text classification \cite{raffel2020exploring}, and machine translation \cite{conneau2019cross}. The transformer architecture uses self-attention mechanism \cite{vaswani2017attention} and it is highly parallelizable on modern Graphical Processing Units (GPUs), providing major benefits over older models like Long Short Term Memories (LSTMs) and Recurrent Neural Networks (RNNs). This has led to fast progress in NLP, with models like BERT exceeding human performance in difficult tasks \cite{wang2018glue} and expanding their application to computer vision, including object recognition and detection \cite{carion2020end}, image classification \cite{dosovitskiy2020image},  and segmentation \cite{guo2022cmt}.\\

Deploying large transformer models on devices with limited resources is challenging due to their high computational and memory requirements. For example,  BERT-Base Transformer  needs $440$ MB of memory and over $176$ Giga FLoating Point Operations (GFLOPs) \cite{fang2022algorithm}  . The computations  are particularly difficult because of the complex attention operations and the quadratic computational complexity related to the length of input sequences \cite{qiu2019blockwise}. Attention operations in transformer models become increasingly dominant as the input sequence length grows. For BERT-Base Transformer deployed on edge platforms, with a sequence length of $512$, the attention operations account for about half of the total execution time, and this figure rises to $70\%$ when the sequence length extends to $768$ \cite{zhou2022energon}. Therefore, finding efficient ways to handle attention operations is crucial for speeding up transformers.\\

Many studies utilized sparsity to mitigate the quadratic time and space complexity issue. Some techniques save computational effort by using  fixed or static sparse attention patterns \cite{beltagy2020longformer} \cite{qiu2019blockwise} \cite{zaheer2020big}, but their performance is limited \cite{10.1145/3530811}, since the sparse pattern in attention is naturally dynamic, and depends only on the input. 
Other techniques focus on dynamic sparsity, meaning there's no fixed pattern for which parts are sparse (zero). For example, $A^3$ \cite{ham20203} uses various approximation methods to skip calculations of near-zero values, aiming to decrease computational demands, but it requires loading all data onto the chip, which doesn't decrease off-chip DRAM access. SpAtten \cite{wang2021spatten} introduces a cascaded token pruning mechanism that gradually eliminates less important tokens to simplify the workload using a Top-K strategy. Despite being tailored for dynamic decision-making in hardware, Top-K requires significant computational cost. Energon \cite{zhou2022energon} uses a mixed-precision, multi-stage filtering technique to mimic the Top-K pruning approach, but it relies on a special unit to handle sparsity. AccelTran \cite{tuli2023acceltran}  prunes values in all transformer matrix multiplications if they fall below certain predetermined threshold.  $A^3$, Energon, and AccelTran leverage unstructured sparsity to realize efficient deployment for Transformers which leads to non uniform data access and lower efficiency. It's hard to predict these sparsity patterns, which can slow down performance.\\

Other research efforts have been directed towards the removal of attention heads in transformer models, based on the understanding that not all heads contribute significantly to the transformer's performance. Researchers in \cite{voita2019analyzing} add trainable gating parameters attached to each head and regularized with $L0$ loss. In \cite{Michel2019_wn}, the importance of each attention head is gauged based on its sensitivity to the overall loss. This sensitivity is utilized as an indirect measure to determine the significance of each head. In \cite{zhang2021know}, a novel approach termed 'single-shot meta-pruner' is presented. This method involves training a compact convolutional neural network with the specific purpose of identifying and selecting the attention heads that are crucial for preserving the distribution of attention within the model. All of these studies perform pruning at compile time not run time, require retraining to recover the accuracy drop. SpAtten \cite{wang2021spatten} also performs a cascaded head pruning at run time using the Top-K approach, wherein the importance of each attention head is derived by aggregating the absolute values of it's attention outputs of the head throughout the layers, resulting in an aggregate score of importance for each head. SpAtten uses a separate unit to perform head Top-K strategy, which also requires significant computational cost. $A^3$, Energon and AccelTran do not support head pruning.

To overcome the above challenges, we propose integer-based Hybrid Dynamic Pruning (HDP), an algorithm-architecture co-design to enable efficient attention inference. This method operates on multiple levels within the attention matrix, reducing complexity by concentrating on  blocks, and heads. This method employs fine-grained block pruning for removing smaller, less critical blocks and incorporates head pruning to selectively eliminate less important heads, all based on the integer parts of inputs for decision-making. The pruning is done dynamically;  applied during inference without relying on fixed patterns for pruning, and without fine-tuning or retraining. Our main contributions are as follows.

\begin{enumerate}
    \item We propose integer-based, fine-grained block pruning, which  prunes unimportant small~sized blocks, with the observation that self-attention primarily relies on a few critical query-key pairs, highlighting significant redundancy in the self attention mechanism. HDP utilizes the integer part of the input to identify and prune less important blocks, focusing subsequent operations only on the unpruned blocks for enhanced efficiency. 
\item We introduce an early head pruning strategy that identifies and eliminates less important heads based on the integer parts of the input at the initial stages of computation, unlike the method in \cite{wang2021spatten} which performs pruning after all computations.
\item We approximate attention calculation by breaking down the multiplication of Q and K into three components: $integer Q \times integer K$, $integer Q \times fractional K$, and $fractional Q \times integer K$. By summing these components together, we not only approximate the attention outcome but also achieve near-zero pruning, as the $fractional Q - fractional K$ multiplication is omitted. This method effectively reduces computational complexity while maintaining model accuracy during inference.
\item We design and implement an ASIC-based architecture to efficiently execute HDP, utilizing encoder-only models to reduce the critical path to the half and enhance throughput and hardware utilization. Our architecture functions as a co-processor, compatible with existing neural network accelerators. Through carefully designed pipelines and architectural optimizations, we aim to significantly boost performance and reduce energy consumption.
\end{enumerate}
The article is structured as follows: Section II provides background information on transformer acceleration. Section III details the methodology of the HDP framework. Section IV describes the hardware architecture of HDP co-processor. Section V outlines the experimental setup, the baselines used for comparison and discusses the results. Finally Section VI concludes the article.

\section{Background and Motivation}

\begin{enumerate} [label = \Alph*.]
\item \textit{Transformer Algorithm }\\
\begin{figure}
\centering
\includegraphics[width=80mm,height=70mm]{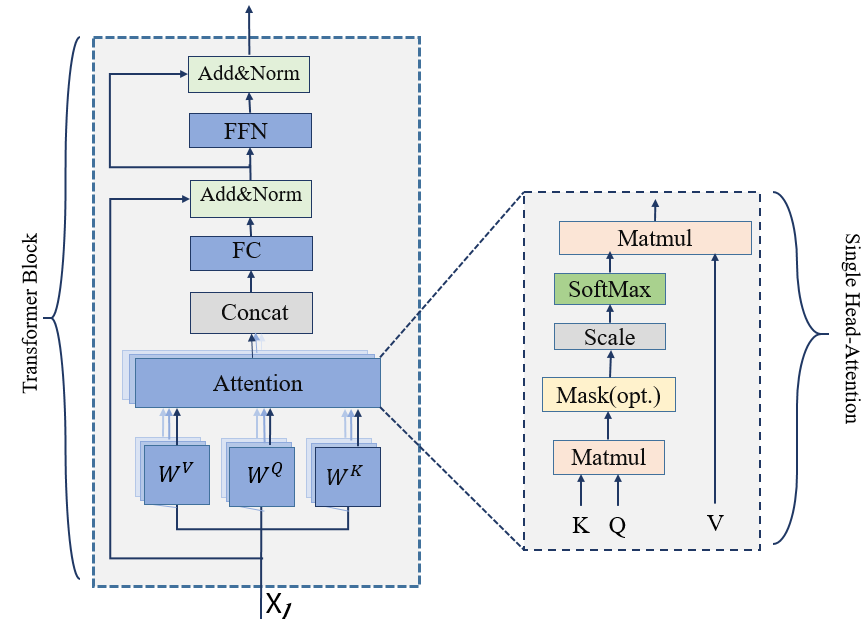}
\captionsetup{justification=centering}

\caption{Transformer Block}
\label{TransformerBlock_SingleHeadAtten}
\end{figure}
Transformers have shown state-of-the-art performance in natural language processing field since their initial introduction in 2017 \cite{vaswani2017attention}. They are often regarded as alternatives to classic CNNs and RNNs  in various situations in real-world  due to their exceptional efficiency and generality. Transformers consist of two essential components: the encoder and the decoder, which are formed by layering many transformer blocks. As depicted in Fig. \ref{TransformerBlock_SingleHeadAtten}, a block comprises three primary elements: linear layers, multi-head self-attention, and the normalization layer. The linear layers include the fully connected Layer (FC), Feed Forward (FFN), and the projection layer. The processes starts with transforming the input vectors into embedding vectors, which are then sequentially passed through a series of processing blocks. Within each block, the input is projected through the projection layer to $Query~(Q)$, $Key~(K)$ and $Value~(V)$ features using the weights $W^Q, W^K$ and $W^V$. Following this, attention mechanisms are utilized on these features to comprehend the long-term relational dependencies present in the input sequence.\\
The attention mechanism, as shown in Algorithm \ref{attentionAlgorithm}, splits the $Q,~K~$and $V$ matrices into multiple smaller sets $Q_h$, $K_h$, and $V_h$ equivalent to the number of  heads $H$. Each of these sets forms an "attention head", see Fig. \ref{TransformerBlock_SingleHeadAtten} right side. Inside each head the output is computed as follows:
$$\text{Attention}(Q_h, K_h, V_h) = \text{softmax}\left(\frac{Q_h{K_h}^T}{\sqrt{d_k}}\right)V_h. $$
The process starts with the computation of the dot product between $Q_h$ and $K_h$, followed by scaling. This step computes the alignment or similarity between each pair of tokens, the fundamental components of a sequence. The resulting matrix represents each token's relationship with every other token. Then a row-wise softmax is applied to obtain the attention~probability. When processing a particular token, the attention~probability weights tell the model how much attention to give to each token in the sequence. After that, the attention~probability is multiplied with the Value vector $V_h$, resulting in a weighted sum that forms the output of the single head attention layer, the resulting output is a combination of information from all tokens, with each token's importance and relevance determined and weighted by the attention mechanism. Each head independently calculates attention result. The results are concatenated to get the end result. The concept entails that each head possesses the ability to concentrate on distinct segments of the input sequence, or capture different types of relations within the data. \\
The attention mechanism fulfills various functions: it enables the model to focus  on the most significant tokens of the input sequence, captures long-term relationships within the sequence, and greatly enhances the model's awareness of the context. This selective attention and context-aware processing are critical for complicated tasks such as language translation \cite{10255243}, text summarization \cite{9464764}, and  other NLP tasks.

\RestyleAlgo{ruled}

\begin{algorithm}[t]
\caption{Attention }\label{attentionAlgorithm}

\SetAlgoLined\SetArgSty{}

\LinesNumbered

\ShowLn \KwIn{ $\{ K,Q,V \} \in \mathbb{R}^{ l \times d}$ }
$l$: sequence length, $d$: input feature dimension\\
Number of heads: H;\\
$Q_h, K_h, V_h \leftarrow$ Split $Q,K,V$ into H chunks;\\
$d_h=d/H$;\\
\For { $head_{id}$~$\leftarrow$~$0$~\KwTo~H}
{
    $attention\_score = {Q_h {K_h}^{T}}$ \;
    $attention\_score \in \mathbb{R}^{l \times l}$ \;
    $attention\_score = \frac{attention\_score}{\sqrt{d_h}}$ \;
    \For{$row_{id}$~$\leftarrow$~$0$~\KwTo~$l$}
    {
        $attention\_prob = \text{softmax}(attention\_score[row_{id}])$ \;
    }
    $result[head_{id}] = attention\_prob \cdot V_h$ \;
}
$attention\_out = \text{concat}(result[head_{id}])$ \;   

\ShowLn\KwOut{ $attention\_out \in \mathbb{R}^{ l \times d}$ \; }
\end{algorithm}

In addition to the multi-head attention component, transformer models also employ other processes such as normalization and FFN. The FFN in a transformer is composed of two FC layers. The function of the FFN can be mathematically expressed as:
$$\text{FFN}(X) = \text{GELU}(XW_1 + b_1)W_2 + b_2,$$
where $X, W_1, b_1, W_2,$ and $b_2$ represent the input to the FFN, the weight matrices, and the bias vectors, respectively. The Gaussian Error Linear Unit (GELU), serves as a special activation function, providing non-linear transformation to the output of the first layer before it is passed on to the second layer \cite{hendrycks2016gaussian}, where it is widely used in Transformers. Our focus in this work is the attention layer as it is the bottleneck for Transformers models. \\

\item \textit{Complexity of Attention Layers }\\
The primary computational requirements of attention layers in a neural network model are pairwise dot-products performed on a set of $l$ vectors. As a consequence, the complexity is $O(l^2d)$. The computational cost associated with attention operations escalates as the length of the input sequence $l$ increases, while $d$ is usually fixed. By measuring the time consumed in attention layer for Bert-base model \cite{devlin2018bert}, on both embedded GPU and CPU platforms, it's observed that attention operations consume half of the total computational time for input sequences longer than $512$. Furthermore, these operations consume almost $70\%$ of computational time for input sequences reach $768$ \cite{zhou2022energon}. Therefore, in scenarios where processing long input sequences is necessary, attention operations emerge as the primary computational bottleneck. Also because of the advancements in linear layers through weight pruning \cite{han2015deep}, quantization \cite{zhou2018adaptive}, and specialized accelerators \cite{chen2014diannao} \cite{chen2016eyeriss}, there is a need to optimize the attention mechanism to ensure balanced computational efficiency in transformer models.\\
\item \textit{Dynamic sparsity in Multi Head Self Attention}\\
Sparse attention methods are driven by the recognition that not all attention probabilities are significant. After applying a row-wise $Softmax$ function to get the $attention\_prob$, most likely  a small subset of the scores in each row will impact the $attention\_out$ results. Similarly, sparse multi-head attention strategies are typically motivated by the recognition that not all heads in the attention mechanism have an equal impact on the final output. This suggests that some heads in the attention model might have a minimal impact on the overall outcome. These observations point to the existence of redundant heads and insignificant attention probabilities in the attention architecture, indicating that the model’s performance could be enhanced by focusing on the most crucial heads and removing unimportant $attention\_scores$.\\
By examining the attention weights matrix ( $attention~probability$) generated from different heads across various layers and inputs, we can say that the significance of individual heads in a transformer's multi-head attention mechanism depends only on the input data. Varying not only with the input data but also with their position within the model's layered architecture. Fig. \ref{Visualization of attention} presents the attention weights of various heads across different layers and inputs in a BERT-base model fine tuned on the SST-2 dataset \cite{socher2013parsing}. As illustrated in Fig. \ref{attention weights visualization for Input1}, there is notable variation in the attention weights for the same head across different layers. For example, the attention values for the eleventh head, highlighted by the red box, show considerable fluctuation across Layers 9, 10, and 11. Additionally, the attention patterns of the same head within the same layer also exhibit significant differences when compared across various inputs. This is clearly depicted in Fig. \ref{attention weights visualization for Input1} and Fig. \ref{attention weights visualization for Input2}, where the green boxes depict the differences in attention values for the same heads and layers when subjected to different inputs. Specifically, Head 0 and Head 1 in Layer 11 exhibit notably low attention weights for Input 1. In contrast, for Input 2, Head 1 and Head 2 in Layer 11 display significantly higher attention values, highlighting the data-dependent nature of attention mechanisms in the model. Furthermore,  it is clear for most of the heads that it is only a subset of the attention weights which have high magnitude, where there is no fixed pattern for the important weights. \\

\begin{figure}
     \centering
     \begin{subfigure}[b]{0.5\textwidth}
         \centering
         \includegraphics[width=\textwidth]{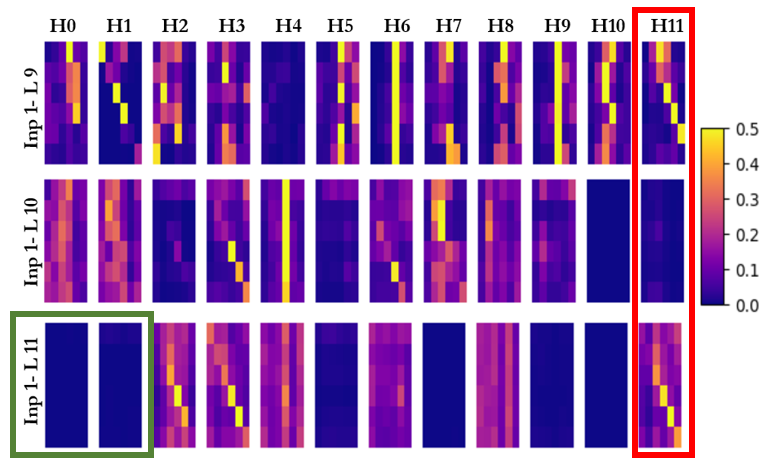}
         \caption{}
         \label{attention weights visualization for Input1}
     \end{subfigure}
     \hfill
     \begin{subfigure}[b]{0.5\textwidth}
         \centering
         \includegraphics[width=\textwidth]{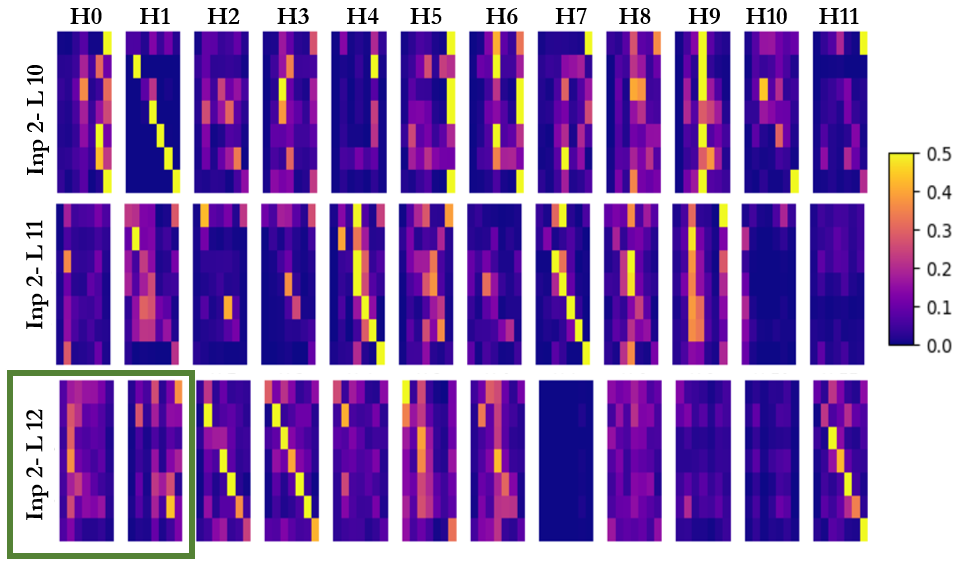}
         \caption{}
         \label{attention weights visualization for Input2}
     \end{subfigure}
    \caption{Attention Probability Analysis in BERT-Base Model for (a) Input 1, (b) Input 2. The red box indicates the variability of attention probabilities across layers for the same input, for a single head (Head11) across different layers (Layers 9, 10, and 11). The green boxes highlight the contrast in the attention probabilities for the same head and layer with two distinct inputs. Head 0 and Head 1 in Layer 11 show lower values for Input 1, while Head 1 and Head 2 in the same layer exhibit significantly higher values for Input 2.}
    \label{Visualization of attention}
\end{figure}
This variability highlights the complex, dynamic nature of how transformers process information. For a given input, certain heads in specific layers may become more active, focusing on particular aspects of the data. This activation can differ from one input to another, reflecting the adaptive response of each head to the unique characteristics of the data it encounters. Similarly, the role of a head may differ across layers. Furthermore,  the high magnitude attention weights subset is dynamic depending on the input sequence , and different heads have different subsets. This data-dependent behavior of attention heads is a critical consideration in model analysis and optimization and motivates us to explore effective methods to eliminate unimportant heads and query-key relations and save computations.\\

\end{enumerate}

\section{Algorithmic Optimization}
In this section, we discuss the algorithmic optimizations that enhance the transformer model's efficiency and performance, focusing on block, head pruning and approximation techniques. These optimizations are key to reducing computational complexity and memory access.\\
\begin{enumerate}[label = \Alph*.]
\item \textit{Block Pruning}\\
For the attention score matrix, most of the query-key relations are not important and can be pruned safely, many methods have been used to prune these relations. Top-$K$ pruning method \cite{wang2021spatten} is used to prune the attention weights where a whole row can be pruned, but this requires a retraining to recover the accuracy, also it requires a specialized hardware to get the $k$ most significant attention weights. Energon \cite{zhou2022energon} avoided the Top-K selection and used the mean filtering as a practical approximation instead, but it still has a separate unit to perform this operation, also faces data duplication overhead due to the multi-round filtering method employed. In Energon and AccelTran \cite{tuli2023acceltran} the pruning is done in an element-wise pattern which results in an irregular sparse matrix, where zeros are randomly spread over the matrix and this results in an irregular memory  access and stalls in the hardware.\\
To address  these challenges we propose integer-based block pruning, where pruning decision is exclusively determined by  the integer parts of the numbers. We employ a small block size for pruning to eliminate the necessity for retraining and to guarantee a more organized and hardware-compatible sparsity pattern.\\
 Initially, multiplication is conducted only on the integer parts of $Q$ and $K$ to obtain $Integer\_atten$. For each $2\times2$ block, we calculate its importance, $\theta$, as the absolute sum of the values within the block. For each row of blocks, we determine the block pruning ratio, $\Theta$, using a method similar to that in Energon, which involves calculating the minimum, maximum, and mean importance values, along with a predefined pruning ratio, $\rho_B$, as shown in Algorithm \ref{BlockHeadPruningApproximation } line \ref{equation}. Blocks with importance, $\theta$, falls below the row-specific threshold, $\Theta$, are pruned and the mask value for the block is assigned to $0$. When a block is pruned, subsequent computations for that block are omitted. Conversely, if a block is retained (mask is 1), the final attention result is approximated, with the approximation technique detailed in \ref{approxItem} The block pruning mechanism is expressed in details in Algorithm \ref{BlockHeadPruningApproximation } lines \ref{startblockPruning} to \ref{endblockPruning} and in Fig. \ref{BlockPruningHeadPruningApproximation}.\\ 

\begin{figure*}
\centering
\includegraphics[width=170mm,height=85mm]{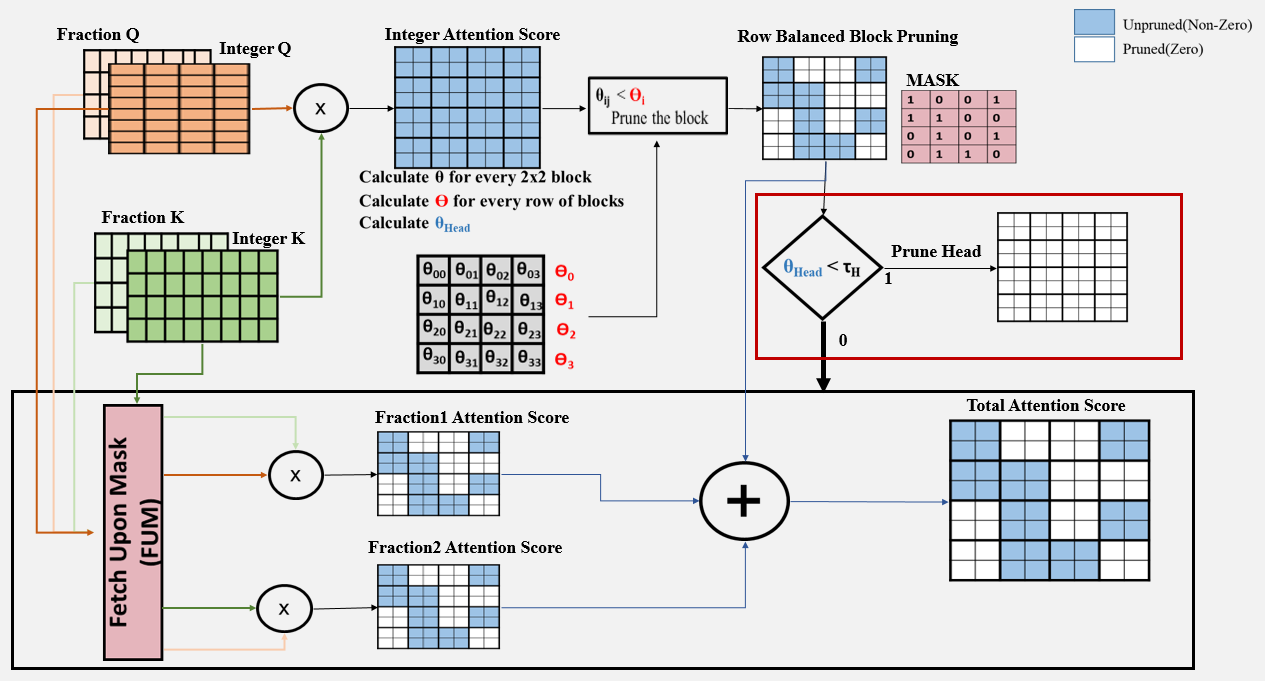}

\caption{Block Pruning, Head Pruning, and Approximation in Integer-Based Row-Balanced Block Sparsity: Pruning is applied to the result of $Integer Q \times Integer K$ based on the comparison between the importance $\theta$ of each block in a row and the row's threshold $\Theta$. Head pruning, highlighted by a red box in the image, is conducted for heads with $\theta_{Head}$ below the predefined threshold $\tau_H$. The approximation process, highlighted by a black box in the image, involves generating fractional components and adding them to the pruned integer results to obtain the final output. This image represents an example with $Q$ of size $8 \times 4$, $K$ of size $4 \times 8$, and a block size of $2 \times 2$.}
\label{BlockPruningHeadPruningApproximation}
\end{figure*}

\item\textit{Approximation}\label{approxItem}\\
For blocks with mask value equals $1$, $\theta$ exceeds $\Theta$, we proceed to approximate the final attention outcome after obtaining the $Integer\_att$ result. This involves calculating two fractions: the product of $Q$'s fractional component with $K$'s integer component, and vice versa, yielding $Frac1\_atten$ and $Frac2\_atten$, respectively. The ultimate attention score is obtained by summing these two fractions with the $Integer\_att$.  This method not only approximates the attention outcome but also enables near-zero pruning, which has minimal impact on model accuracy during inference \cite{jietal2021on}. Specifically, when two numbers are close to zero, their integer parts are zero, leading all three components to be zero. Consequently, the multiplication of fractional parts is omitted, resulting in effective near-zero pruning. This approximation process is highlighted within the black box in Fig. \ref{BlockPruningHeadPruningApproximation} and in Algorithm \ref{BlockHeadPruningApproximation } lines \ref{start approx} to \ref{end approx}.\\

\item\textit{Head Sparsity}\\
Not all heads are essential, and many can be pruned without affecting the overall performance. Unlike the method in \cite{wang2021spatten}, where head importance is assessed after completing all computations for the head, we introduce an early head pruning approach. To evaluate head importance, $\theta_{Head}$, we compute the absolute summation of all values in $Integer\_att$. Heads with $\theta_{Head}$ below a predefined threshold $\tau_H$ are completely pruned and the rest of computations of this head are skipped. The threshold $\tau_H$ is a parameter that will be profiled. The head pruning process is visually indicated by the red box in Fig \ref{BlockPruningHeadPruningApproximation} and described in Algorithm \ref{BlockHeadPruningApproximation } at line \ref{HeadPruningAlgo}.\\

\end{enumerate}

\RestyleAlgo{ruled}

\begin{algorithm}
\caption{Block, Head Pruning and Approximation (One Head)}\label{BlockHeadPruningApproximation }
\SetAlgoLined\SetArgSty{}
\LinesNumbered
\ShowLn \KwIn{Quantized $\{ K,Q,V \} \in \mathbb{R}^{ l \times d}$ }
Block size: $2 \times 2$\\
Block Pruning Ratio: $\rho_B$\\
Head Pruning Threshold: $\tau_H$\\
Integer part of Q: $IQ$, Fractional part of Q: $FQ$\\
Integer part of K: $IK$, Fractional part of K: $FK$\\

$Integer\_atten \leftarrow IQ \cdot IK^{T}$ \label{startblockPruning} \;
\tcc{ For the Integer\_atten matrix with 2x2 block size, iterate over the $l/2$ rows of blocks}
\For{$i$~$\leftarrow$~$0$~\KwTo~$l/2$}
{   
    \tcc{Process each 2x2 block within the row}
    \For{$j$~$\leftarrow$~$0$~\KwTo~$l/2$} 
    {
        $\theta_j \leftarrow \sum\limits_{x \in block_j} |x|$ \;
        $\theta_{Head} \leftarrow \theta_{Head} + \theta_j$ \;
    }
    $min_i \leftarrow \min(\theta_j)\quad\forall\quad\theta_j$ \;
    $max_i \leftarrow \max(\theta_j)\quad\forall\quad\theta_j$ \;
    $mean_i \leftarrow \sum\limits_{0}^{l/2} \theta_j \Bigg / (l/2)$ \;

    \[
    \Theta_i = 
    \begin{cases}
        \rho_B \times max_i + (1 - \rho_B) \times mean_i & \text{if } 0 \leq \rho_B < 1 \\
        -\rho_B \times min_i + (1 + \rho_B) \times mean_i & \text{if } -1 < \rho_B < 0
    \end{cases}
    \] \label{equation}

    $Mask_j^i \leftarrow (\theta_j < \Theta_i) ? 0 : 1$ \;
    $Integer\_atten \leftarrow Integer\_atten \odot Mask_j^i$ \label{endblockPruning} \;
}

\eIf{$\theta_{Head} > \tau_H$ \label{start approx}}
{
    \For{$i$~$\leftarrow$~$0$~\KwTo~$l/2$}
    {
        \For{$j$~$\leftarrow$~$0$~\KwTo~$l/2$}
        {
            \If{$Mask_j^i == 1$}
            {
                $Frac1\_atten_j^i \leftarrow IQ_i \cdot FK_j^{T}$ \;
                $Frac2\_atten_j^i \leftarrow FQ_i \cdot IK_j^{T}$ \;
            }
        }
    }
    \tcc{approximated attention score }
    $approximation \leftarrow Integer\_atten + Frac1\_atten + Frac2\_atten$ \label{end approx} \;
    $attention\_score \leftarrow approximation / \sqrt{d}$ \;
    $attention\_prob \leftarrow softmax(attention\_score)$ \; 
    $result \leftarrow attention\_prob \cdot V$ \;
}
{
    \tcc{Prune the whole head }
    $result$ = $0$; \label{HeadPruningAlgo}\\
}
\ShowLn\KwOut{ $result$ $\in  \mathbb{R}^{ l \times d}$\; }
\end{algorithm}


\newcommand{\cir}[1]{\tikz[baseline]{%
    \node[anchor=base, draw, circle, inner sep=0, minimum width=1.2em]{#1};}}
    
\section{Hardware Architecture}
\begin{figure}

\includegraphics[width=90mm,height=40mm]{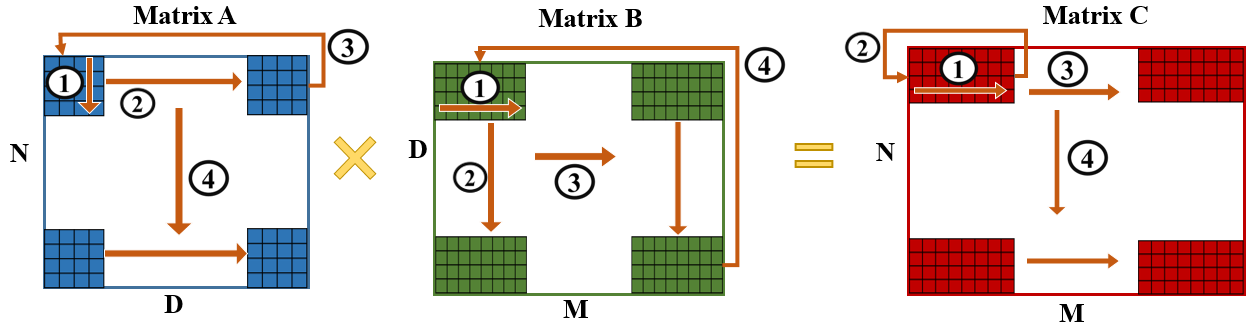}
\color{blue}
\begin{footnotesize}
            \begin{minipage}{\textwidth}
               {\fontfamily{Courier}\selectfont

                \begin{Verbatim}[commandchars=\\\{\}]
                
\textbf{for (int i=0; i < N; i=i+4)} \textcolor{ForestGreen}{//Tiling along N dimension}
 \textbf{for(int j=0; j < M; j=j+8)} \textcolor{ForestGreen}{//Tiling along M dimension}
  \textbf{for(int k=0; k < D; k=k+4)} \textcolor{ForestGreen}{//Output Stationary}
   \textbf{for (int ii=0; ii < i+4; ii++)} \textcolor{ForestGreen}{//A input stationary}
    \textbf{for(int jj=0; jj < j+8; jj++)}  \textcolor{ForestGreen}{//B input reuse}
     \textbf{for(int kk=0; kk < k+4; kk++)} \textcolor{ForestGreen}{//Partial sum reuse}
      \textbf{C[ii][jj] += A[ii][kk]*B[kk][jj]} \textcolor{ForestGreen}{//MAC_OPERATION}

                \end{Verbatim}
                
                }
            \end{minipage}
            \end{footnotesize}
\caption{Matrix Multiplication Tiling with the Dataflow.}
\label{Tiling}
\end{figure}

Current attention accelerators and traditional CPUs/GPUs lack the capability to execute the proposed hybrid dynamic sparse attention technique efficiently. To address this gap, we are introducing a novel HDP accelerator. HDP is developed to function as a co-processor and is compatible with a variety of neural network accelerators for easy integration.\\
\begin{enumerate}[label = \Alph*.]
    \item \textit{Architecture Overview}\\
The HDP architecture, depicted in Fig.  \ref{CoresArchitectureOverview}, compromises multiple cores, the architecture of individual core is shown in Fig. \ref{CoresArchitectureOverview} middle part. Each core is composed of an array of  processing elements  (PE Array), a Sparsity Engine (SE), an adder, and a softmax unit. The PE Array handles matrix multiplication tasks such as $Q \times K^T$ and $attention\_prob \times V$,  also calculating importance values. The SE is tasked with identifying which blocks to prune and deciding whether a head should be pruned or not.\\
     
\begin{figure*}

\includegraphics[width=195mm,height=60mm]{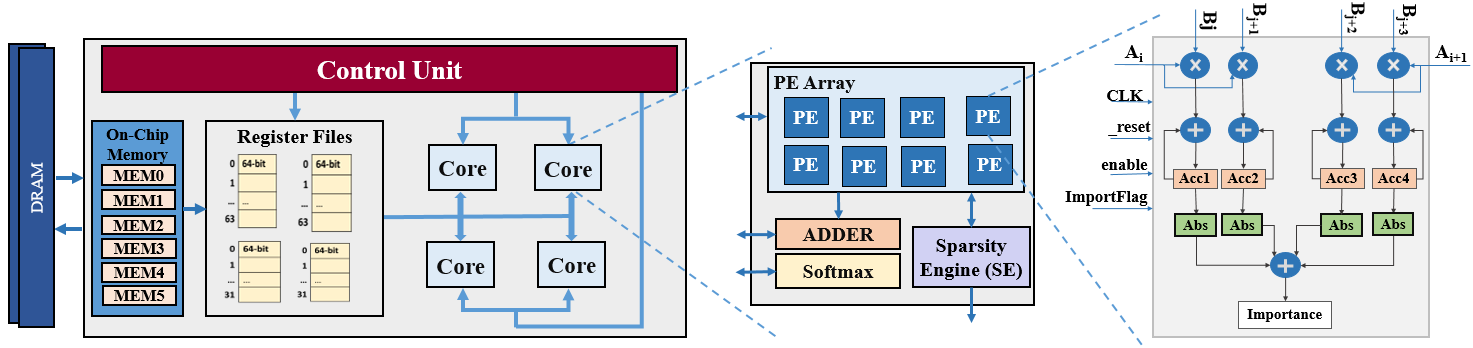}

\caption{HDP Architecture Overview.}
\label{CoresArchitectureOverview}
\end{figure*}
     


     \textit{Workflow:} Once $Q,K,$ and $V$  are generated and quantized by another processor in fixed point 16 bit format and stored in memory, HDP processes each attention head sequentially.  It employs tiled matrix multiplication for these operations. Initially, the integer components of $Q$ and $K$  are retrieved from off-chip memory into on-chip memory for the computation of $Integer\_Q \times Integer\_K$ using tiling.  The SE then uses the computed importance values for each block to create a mask indicating which blocks are not pruned. This approach prevents unnecessary data fetching for pruned blocks, reducing memory access and computational overhead. Once $Integer\_Q \times Integer\_K$ computation is complete, and the head importance is assessed, the SE decides whether a head will be pruned. If so, the remaining computations for that head will be skipped and proceeds to the next head. For heads that remain unpruned, a Fetch Upon Mask (FUM) strategy is utilized.  If the mask value is 0, indicating a pruned block, the corresponding  K values will not be fetched, and the computation for that block is skipped. If the mask value is 1, the corresponding Q and K values are fetched, and the processing element (PE) array calculates the two remaining fractions ($Integer\_Q \times Frac\_K$ and $Frac\_Q \times Integer\_K$) simultaneously. These results, along with the integer results from the previous step, will be added together using an ADDER module to obtain the total $attention\_score$. After performing all $Q \times K$, a row-wise softmax will be applied to the $attention\_score$. Then, the PEs will be utilized to perform the calculation $attention\_score \times v$. Specifically, the first and second PEs in the first row calculate $Integer\_attention\_score \times Integer\_v$, while the third and fourth PEs in the first row compute $Integer\_attention\_score \times Frac\_v$. The first and second PEs in the second row calculate $Frac\_attention\_score \times Integer\_v$, and the third and fourth PEs in the second row calculate $Frac\_attention\_score \times Frac\_v$. Finally, all these results will be summed using the ADDER module to obtain the final output. The attention results for each tile will be immediately stored in DRAM memory upon completion. Consequently, host DNN accelerators can access these results to perform subsequent computations. In the sections that follow, we provide a more in-depth exploration of each module and the optimizations we have proposed to enhance their functionality.\\


\item Tiling and Dataflow\\
A significant portion of the computational workload in transformer models is attributed to matrix multiplication, necessitating optimization to boost accelerator performance. We propose the implementation of tiled matrix multiplication, a technique primarily employed in GPUs \cite{niu2022tilespgemm}, to enhance the efficiency of these operations in our accelerator design. Tiling enhances resource efficiency and enables parallel computation, as depicted in Fig. \ref{Tiling}. The first $4 \times 4$ tile from matrix $A$ is multiplied by the first $4 \times 8$ tile from matrix $B$, with the partial results stored in a $4 \times 8$ tile in matrix $C$, denoted by \cir{1} in all matrices. Subsequently, the process advances along \cir{2} in the tiles of $A$ and $B$ to accumulate additional partial sums in $C$. During this stage, an output stationary dataflow approach is employed, facilitating the reuse of partial sum outputs in the accumulator. Additionally, a local $A$ stationary strategy is implemented, meaning that while outputs are reused in the outer loop, inputs from matrix $A$ are retained and reused in the inner loop \cite{venkatesan2019magnet}. 
Next, the process proceeds along \cir{3} in all matrices, followed by a moving along \cir{4}.  \\
\item Processing Elements\\
The processing element, shown in Fig.  \ref{CoresArchitectureOverview} (right), serves as a fundamental computational unit within the accelerator handling all matrix multiplication operations. Operating in an output stationary mode, and behaving similar to systolic array PE; it receives rows from tiles of the first matrix and columns from tiles of the second matrix as inputs, one input at a time. It multiplies these values and stores the intermediate sums in accumulators until the entire row from the first matrix has been multiplied by the corresponding column from the second matrix. At this point, the accumulators hold the final results for the tile of the result matrix.  In the case of $Integer~Q \times Integer~K$ multiplication, these results are also utilized to determine the block's importance,  as the output from a processing element corresponds to a block in the result matrix. The importance of the block, as illustrated in Fig. \ref{CoresArchitectureOverview}, is equal to the absolute sum of the accumulators.\\
\item Sparsity Engine\\
The Sparsity Engine  is responsible for determining the sparsity pattern at both block and head levels. Illustrated in Fig. \ref{SparsityEngine}, the Sparsity Engine's internal architecture takes in importance scores from the PE and stores them in its internal memory. Additionally, it keeps track of the minimum, maximum, and total sum of these importance values for every row of blocks. Upon receiving the $END\_R$, which signals the completion of a full row in the result matrix or, equivalently, the multiplication of a row from the first matrix by all columns in the second matrix in  $Integer~Q \times Integer~K$ multiplication, the engine calculates the block pruning threshold $\Theta$ for that specific row. This calculation is based on the equation provided in line \ref{equation} of Algorithm \ref{BlockHeadPruningApproximation }. Additionally, the engine generates the $Mask$ for the row by subtracting $\Theta$ from the importance values of the blocks. If the result is negative, the block falls below $\Theta$ and is marked for pruning.\\
Furthermore, when the $END\_H$ is received, signaling the completion of the $Integer~Q \times Integer~K$ multiplication, the engine utilizes the $\theta_{Head}$ value it has computed representing the total sum of all importance values across the entire head. Upon receiving this flag, the engine compares it with $\tau_H$, an input parameter denoting the head pruning threshold. If $\theta_{Head}$ falls below $\tau_H$, the head is assumed redundant and is thus excluded, allowing the accelerator to bypass any further calculations for this head and proceed to processing the subsequent head.\\

\item Softmax Module\\
Once the attention scores are obtained, a row-wise softmax function is applied defined as $e^{s_i} / \sum_{j} e^{s_j}$. For every input received to the module, the exponent is approximated using $2^{nd}$ order polynomial. The exponent results are stored in internal memory, and the sum of these exponent results is calculated. By the end of every row, the reciprocal of the sum is computed using a linear approximation. Then the exponent values are multiplied by the reciprocal to generate the softmax result.\\



\end{enumerate}

\begin{figure}
\includegraphics[width=95mm,height=80mm]{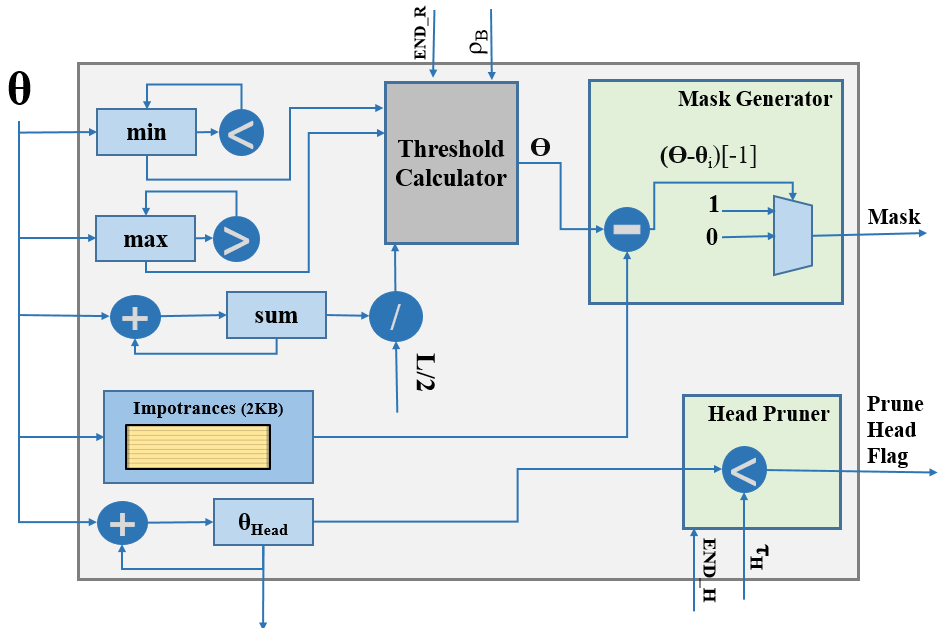}

\caption{Internal architecture of Sparsity Engine (SE).}
\label{SparsityEngine}
\end{figure}
\section{Evaluation}
\subsection{Algorithm Evaluation}
\subsubsection{Evaluation Models and Datasets}
To validate the efficacy of our proposed method, HDP, we focused on evaluating encoder-only models. For this purpose, we utilized BERT-Tiny  \cite{turc2019well} and BERT-Base  \cite{devlin2018bert}, two well-established pre-trained models. BERT-Tiny consists of two encoder layers, each with a hidden dimension of 128 and two attention heads, while BERT-Base contains 12 encoder layers, each with a hidden dimension of 768 and 12 attention heads. These encoder-only models are particularly promising for applications in areas such as machine translation \cite{zhu2020incorporating} and language generation \cite{rothe2020leveraging}, where their efficiency and scalability can be leveraged for improved performance. Our evaluation was conducted on two benchmark tasks: SST-2 and COLA, both sourced from the GLUE benchmark \cite{wang2018glue}.\\
\subsubsection{Dynamic Inference with Transformer}
we present the results of our experiments, encompassing various aspects of our study. These results provide a comprehensive overview of the performance and efficacy of the proposed HDP algorithm, from exploring various block pruning ratios and profiling head thresholds to conducting thorough comparisons with the established Top-k block pruning method.\\
\begin{enumerate}[label=(\alph*)]
    \item Block Pruning: Our baseline for comparison is the Top-k block pruning method with a block size of $2 \times 2$. As depicted in Fig. \ref{TopKVSIMLDP}, the Top-K method can prune up to $75\%$ of all blocks with $1\%$ accuracy loss, whereas HDP can achieve a pruning ratio of $70\%$. However, for pruning ratios exceeding $80\%$, HDP no longer serves as a reliable approximation to the Top-K method. This discrepancy is evident in the figure, where the accuracy of HDP is significantly higher than that of Top-K, indicating that the model is unable to accurately determine the correct block pruning threshold, $\Theta$. This issue arises because the model incorrectly assumes that it is pruning a high percentage of blocks, when in fact it is not, and this attributed to the assumption in that the mean divides the data into equal halves. 

Both models exhibit an initial improvement in accuracy followed by a decline as the level of sparsity increases. This may be linked to the over-parameterization of the BERT model \cite{hao2019visualizing}, and it can be though as in NLP, if unimportant tokens are removed, the model can focus more on the important token resulting in better accuracy.\\
\item Head Pruning:
In HDP, the impact of head pruning is depicted in Fig. \ref{HeadThresholdProfiling}, which demonstrates the threshold profiling and the corresponding accuracies after applying head pruning to the BERT-Base and BERT-Tiny models on the SST2 and CoLA datasets. As anticipated, head pruning is particularly critical for BERT-Tiny, as illustrated in Fig. \ref{tinycola} and Fig. \ref{tinysst}. These figures reveal that less than $2\%$ of the model's heads can be pruned without affecting the accuracy. This sensitivity is due to the model's limited number of heads, with only 4 in total. Consequently, removing even a single head amounts to a significant coarse-grained pruning of one-fourth of all heads, and this is a very large coarse grained pruning to be done without any retraining. On the other hand, head pruning in the BERT-Base model yields more favorable pruning ratios due to its larger number of heads, totaling 144. As depicted in Fig.\ref{baseCola} and Fig.\ref{basesst}, the model can prune approximately $13-17\%$ of its heads with only a $1\%$ decrease in accuracy.\\

\begin{figure}
\includegraphics[width=90mm,height=80mm]{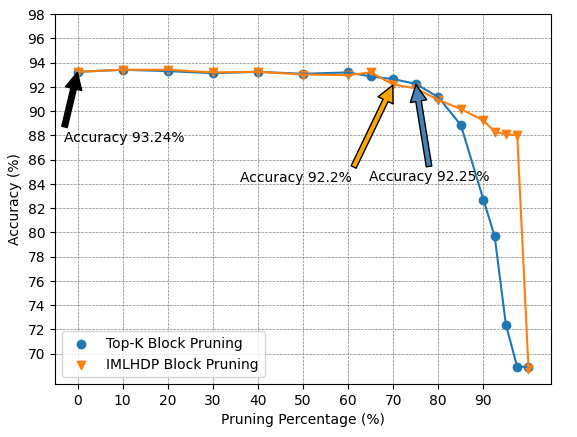}
\centering
\caption{Top-k VS HDP block pruning.}
\label{TopKVSIMLDP}
\end{figure}

\begin{figure*}[ht]
    \centering
    \begin{subfigure}[b]{0.49\linewidth}
        \includegraphics[width=\linewidth]{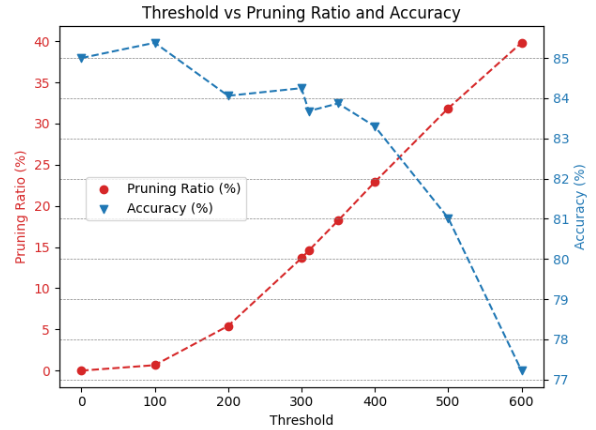}
        \caption{Head Pruning Threshold for BERT-Base on CoLA dataset.}
        \label{baseCola}
    \end{subfigure}
    \hfill
    \begin{subfigure}[b]{.49\linewidth}
        \includegraphics[width=1\linewidth]{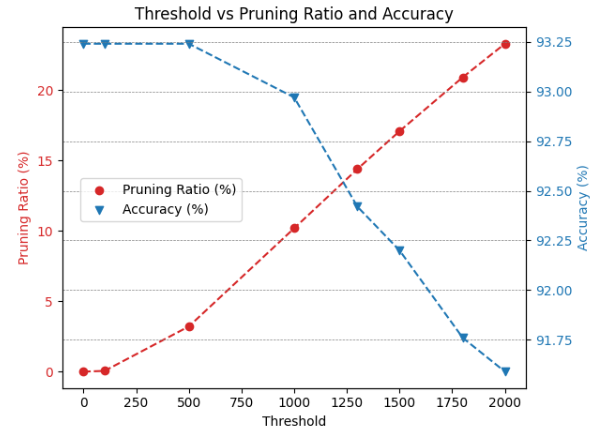}
        \caption{Head Pruning Threshold for BERT-Base on SST2 dataset.}
        \label{basesst}
    \end{subfigure}

    \vspace{1cm} 

    \begin{subfigure}[b]{0.49\linewidth}
        \includegraphics[width=1\linewidth]{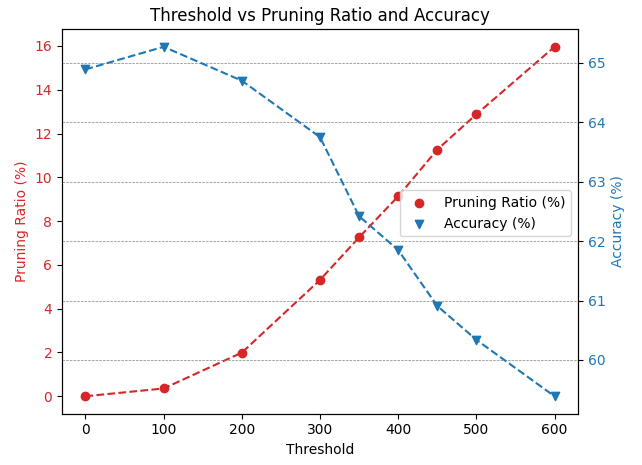}
        \caption{Head Pruning Threshold for BERT-Tiny on CoLA dataset.}
        \label{tinycola}
    \end{subfigure}
    \hfill
    \begin{subfigure}[b]{0.49\linewidth}
        \includegraphics[width=1\linewidth]{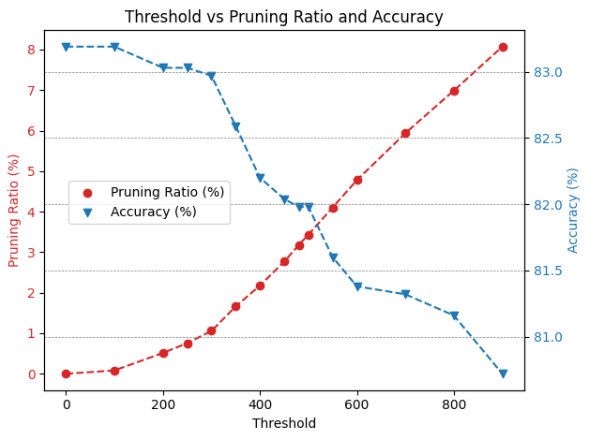}
        \caption{Head Pruning Threshold for BERT-Tiny on SST2 dataset.}
        \label{tinysst}
    \end{subfigure}
    \caption{Head Pruning Threshold for BERT-Base and BERT-Tiny on SST2 and CoLA.}
    \label{HeadThresholdProfiling}
\end{figure*}
\item Approximation:
To assess the effectiveness of the proposed approximation method, we will examine its impact on the models' accuracy. Fig. \ref{ApproximationEffect} illustrates the accuracy of models employing block pruning with and without the approximation technique. 
For the BERT-Base model, as depicted in Fig. \ref{Approxbasesst2} and Fig. \ref{Approxbasecola}, the model's performance remains almost the same, suggesting that the approximation does not negatively affect the model while providing benefits in terms of computational efficiency. In contrast, for the BERT-Tiny model, as shown in Fig. \ref{Approxtinysst} and Fig. \ref{Approxtinycola}, the model is more sensitive to the approximation, experiencing a greater effect on its performance. 
While both BERT-Base and BERT-Tiny have an identical hidden size of 64 for each head, the reduced number of heads in BERT-Tiny amplifies the impact of pruning within a head on its overall performance. The near-zero pruning strategy, which allocates higher softmax values to unpruned elements, allows the model to focus more on crucial $Q-K$ relations, thereby enhancing its concentration on important components. However, in some instances, the approximation may lead to reduced accuracy. This could be attributed to the nature of the approximation itself, as the fraction removed is not uniform across all values. Consequently, this could inadvertently lower the attention score for important $Q-K$ relations in certain scenarios.\\

\item Net Pruning: 
Fig. \ref{NETSparsity} demonstrates the combined effect of block pruning, head pruning, and approximation techniques on the model's overall sparsity. With a $1\%$ reduction in accuracy, the net sparsity achieved by BERT-Base on the SST2 dataset is $75\%$, matching the pruning percentage attained through the Top-K method. For BERT-Base on the CoLA dataset, the net sparsity is $65\%$.\\
By leveraging net sparsity, we were able to attain a pruning ratio comparable to that of the Top-K method. In the Top-K approach, even within an unimportant head, certain blocks remain unpruned due to their significance within that specific head. However, the entire head may not be crucial overall. By implementing head pruning, we successfully removed such unimportant heads, thereby achieving a higher overall pruning ratio.

\begin{figure*}[ht]
    \centering
    \begin{subfigure}[b]{0.49\linewidth}
        \includegraphics[width=\linewidth]{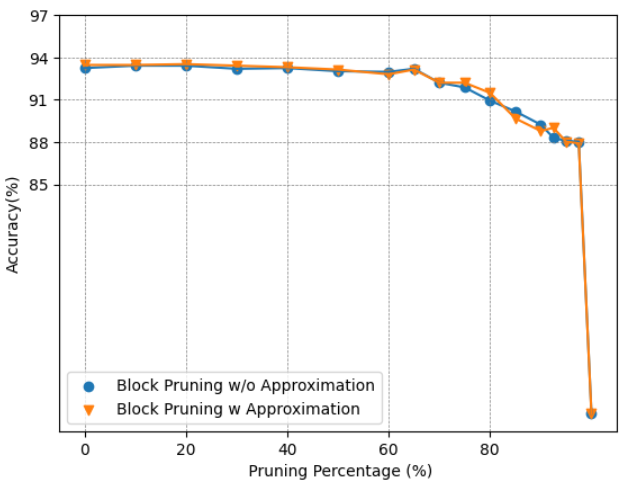}
        \caption{Accuracy of BERT-Base on SST2.}
        \label{Approxbasesst2}
    \end{subfigure}
    \hfill
    \begin{subfigure}[b]{0.49\linewidth}
        \includegraphics[width=\linewidth]{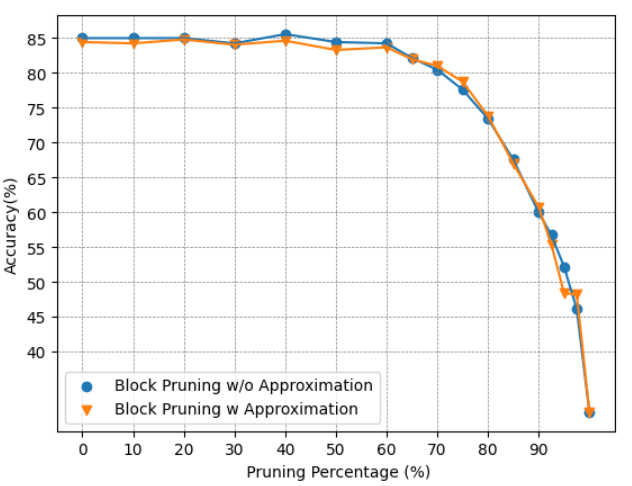}
        \caption{Accuracy of BERT-Base on CoLA.}
        \label{Approxbasecola}
    \end{subfigure}

    \vspace{1cm} 

    \begin{subfigure}[b]{0.49\linewidth}
        \includegraphics[width=\linewidth]{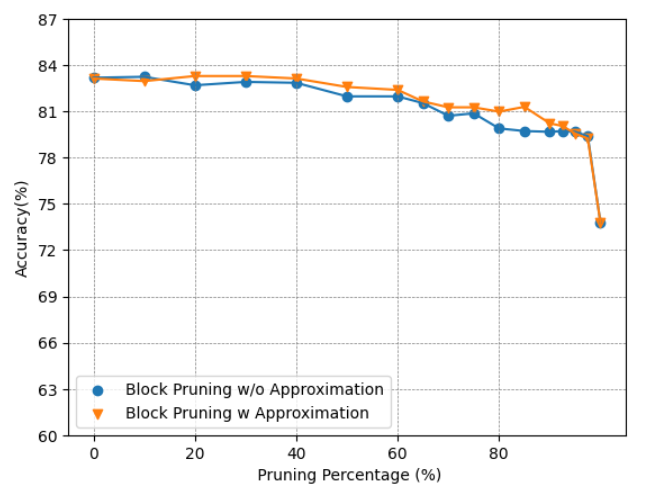}
        \caption{Accuracy of BERT-Tiny on SST2.}
        \label{Approxtinysst}
    \end{subfigure}
    \hfill
    \begin{subfigure}[b]{0.49\linewidth}
        \includegraphics[width=\linewidth]{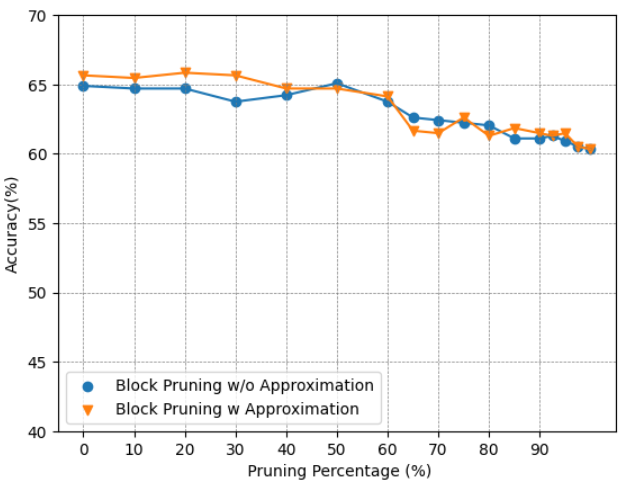}
        \caption{Accuracy of BERT-Tiny on CoLA.}
        \label{Approxtinycola}
    \end{subfigure}
    \caption{Accuracy Comparison of BERT-Base and BERT-Tiny Block Pruning Method with and without Approximation on SST2 and CoLA.}
    \label{ApproximationEffect}
\end{figure*}
\begin{figure}[ht]
    \centering
    \begin{subfigure}[b]{1\linewidth}
        \includegraphics[width=\linewidth]{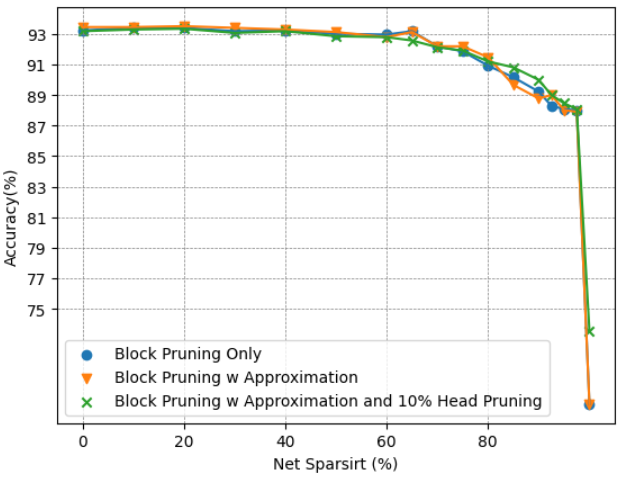}
        \caption{BERT-Base on SST2.}
        \label{netsparsitybasesst2}
    \end{subfigure}
    \hfill
    \begin{subfigure}[b]{1\linewidth}
        \includegraphics[width=\linewidth]{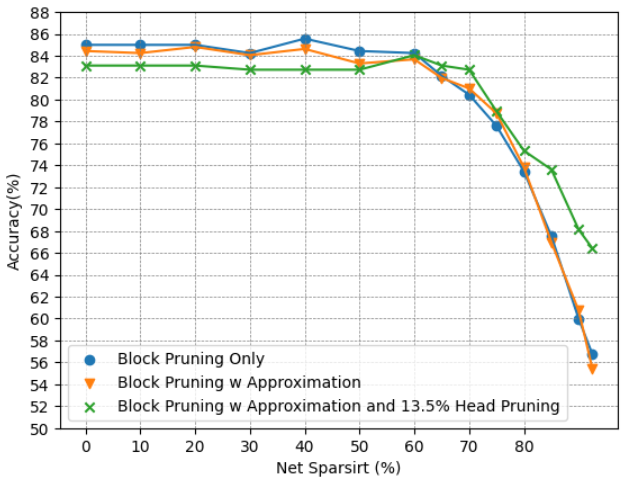}
        \caption{BERT-Base on CoLA.}
        \label{netsparsitybasecola}
    \end{subfigure}


    \caption{Comparison of Accuracy vs. Net Pruning Ratio for BERT-Base  Using Block Pruning, Head Pruning with and with Approximation on SST2 and CoLA Datasets.}
    \label{NETSparsity}

\end{figure}
\end{enumerate}


\subsection{Comparisons with Related Work }
\subsubsection{Comparison with SpAtten}
\begin{figure}[ht]
    \centering
    \begin{subfigure}[b]{1\linewidth}
        \includegraphics[width=\linewidth]{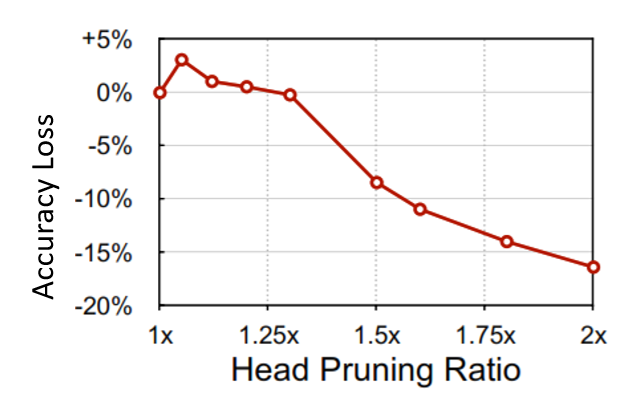}
        \caption{SpAtten Head Pruning Results\cite{wang2021spatten}.}
        \label{SpattenHeadPruning}
    \end{subfigure}
    \hfill
    \begin{subfigure}[b]{.8\linewidth}
        \includegraphics[width=\linewidth]{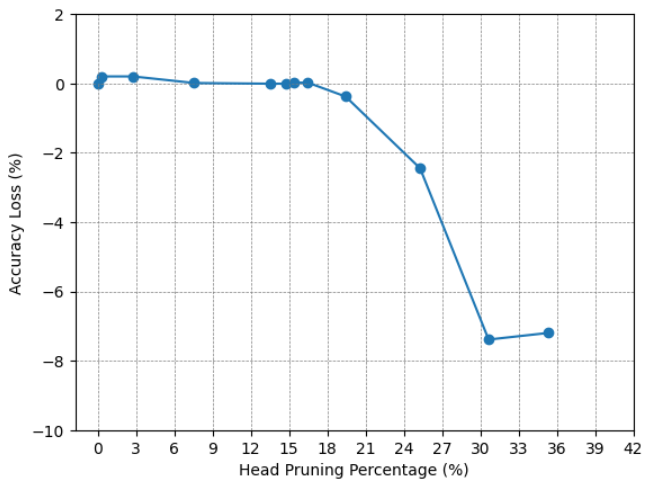}
        \caption{Proposed Fine Tuned Head Pruning Method.}
        \label{basesst}
    \end{subfigure}
      \caption{Accuracy of BERT-Base after dynamic head pruning in (a) SpAtten, (b) Our method fine tuned.}
    \label{HeadpruningComparision}
\end{figure}

To evaluate our proposed head pruning technique, we will conduct a comparison with SpAtten \cite{wang2021spatten}, the only study to date that dynamically applies head pruning directly on hardware platforms. As documented in SpAtten's findings for the BERT-Base model applied to the CoLA dataset, it achieved up to $17\%$ head pruning with no loss in the accuracy. To ensure a fair comparison with SpAtten, we quantized our model to 12 bits and adhered to the identical fine-tuning protocol outlined in their study. The fine-tuning of the BERT-Base model on the CoLA dataset was completed on an average GPU in under two hours, employing various combinations of block pruning ratios and head pruning thresholds without pruning any thing from the first $30\%$ of the layers. Fig. \ref{HeadpruningComparision} shows  these values. After fine-tuning, our method was able to prune approximately $17\%$ of the heads sam as SpAtten. It's worth noting that for higher pruning ratios for example $35\%$ pruning percentage(1.55x pruning ratio), although there is a significant drop in accuracy, the decrease in accuracy is less pronounced in our model which is $7.5\%$ compared to SpAtten which equal $10\%$. SpAtten employs a cascading head pruning approach, where once a head is pruned from one layer, it is also pruned from all subsequent layers. This is in contrast to findings that suggest head importance is only data-dependent, indicating that a head may be important in one layer but not in another, as demonstrated in Fig. \ref{Visualization of attention}.

Table \ref{Comparison of IMLHDP with related works} compares HDP with popular transformer accelerators.

\begin{table}[!hbt]
    \centering
    \caption{Comparison of HDP with related works along different  aspects.}
    \begin{tabular}
    {p{1.75cm}|p{.5cm}|p{.75cm}|p{.79cm}|p{1cm}|p{.75cm}}
    
    \hline
    Work& $A^3$ \cite{ham20203} & SpAtten \cite{wang2021spatten}&Energon \cite{zhou2022energon}&AccelTran \cite{tuli2023acceltran}& \textbf{HDP (Ours)}\\
    \hline 
       Head Pruning &&$\checkmark$&&&$\checkmark$\\ \hline
       Block Pruning &&&&&$\checkmark$\\ \hline
       Approximation &$\checkmark$&&&&$\checkmark$\\ \hline
       Tiled Mat. Mul. &&&&$\checkmark$&$\checkmark$\\ \hline
       Sparsity-aware &&$\checkmark$&$\checkmark$&$\checkmark$&$\checkmark$\\ \hline
       Dynamic Inference&$\checkmark$&$\checkmark$&$\checkmark$&$\checkmark$&$\checkmark$ \\ \hline
       \hline     
    \end{tabular}
    \label{Comparison of IMLHDP with related works}
\end{table}

\section{Conclusion}
In this work we presented HDP, a novel algorithm-architecture co-design to efficiently run dynamic sparse attention models. We first proposed a novel integer-based row-balanced block pruning to prune unimportant blocks in attention matrix and integer based head pruning  to prune unimportant heads. Moreover, we propose an approximation method that reduces the computations and performs near-zero pruning. We also implemented this method in 2 co-processors architecture, HDP-Edge and HDP-Server, to accelerate algorithm on mobile and sever platforms. 


%



\section*{Acknowledgment}
This work was supported by the Khalifa University of Science and Technology under SOCL.

\ifCLASSOPTIONcaptionsoff
  \newpage
\fi

\bibliographystyle{unsrt}
\bibliography{references.bib}
\end{document}